%% file: paper.tex
\documentclass{article}
\usepackage[T1]{fontenc}
\usepackage{lmodern}
\usepackage{graphicx}
\usepackage{booktabs}
\usepackage{siunitx}
\usepackage[margin=1in]{geometry}
\usepackage[authoryear]{natbib}
\usepackage{caption}
\usepackage{subcaption}
\usepackage{hyperref}
\usepackage{setspace}
\title{\includegraphics[height=1em]{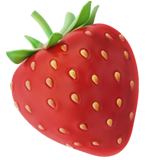}Ichigo: Mixed-Modal Early-Fusion Realtime Voice Assistant}
\author{Alan Dao (Gia Tuan Dao)\textsuperscript{*}, Dinh Bach Vu\textsuperscript{*}, Huy Hoang Ha\textsuperscript{*} \\
\small{\textit{Menlo Research}} \\
\small{\textit{\textsuperscript{*}Equal contribution}} \\
\small{\texttt{alan, bach, rex@menlo.ai}}}
\date{\today}

\begin{document}

\maketitle

\begin{abstract}
    Large Language Models (LLMs) have revolutionized natural language processing, but their application to speech-based tasks remains challenging due to the complexities of integrating audio and text modalities. This paper introduces Ichigo, a mixed-modal model that seamlessly processes interleaved sequences of speech and text. Utilizing a tokenized early-fusion approach, Ichigo quantizes speech into discrete tokens and employs a uniform transformer-based architecture for both speech and text modalities. This method enables joint reasoning and generation across modalities without the need for separate adapters. We present a comprehensive training methodology, including pre-training on multilingual speech recognition datasets and fine-tuning on a curated instruction dataset. Ichigo demonstrates state-of-the-art performance on speech question-answering benchmarks, outperforming existing open-source speech language models and achieving comparable results to cascaded systems. Notably, Ichigo exhibits a latency of just 111 ms to first token generation, significantly lower than current models. Our approach not only advances the field of multimodal AI but also provides a framework for smaller research teams to contribute effectively to open-source speech-language models.
\end{abstract}

\include{1-introduction}

\include{2-model-arch}

\include{3-data}

\include{4-training}

\include{5-results} 

\include{6-conclusion}

\bibliographystyle{plainnat}
\bibliography{references}

\newpage
\appendix
\section{Additional Data and Analysis}

This appendix provides supplementary information on the Audio Speech Recognition (ASR) prompt library and ablation studies conducted during our research.

\subsection{ASR Prompt Library}

Table \ref{tab:asr-prompts} presents a collection of prompts used for the Ichigo Model transcription tasks. These prompts were designed to elicit accurate speech-to-text conversions across various contexts.

\begin{table}[ht]
    \centering
    \caption{Audio Speech Recognition (ASR) Prompt Library for Ichigo Model Transcription Tasks}
    \label{tab:asr-prompts}
    \begin{tabular}{l}
    \toprule
    \textbf{Transcribe Prompts} \\
    \midrule
    Transcribe the following audio clip: <speech> \\
    Convert the spoken words to text: <speech> \\
    What is being said in this audio clip: <speech> \\
    Transcribe the speech in this audio sample: <speech> \\
    Please write down what is being said in the audio clip: <speech> \\
    Generate a transcript from this sound file: <speech> \\
    Recognize the speech in this audio clip: <speech> \\
    Produce a text version of this audio recording: <speech> \\
    \bottomrule
    \end{tabular}
\end{table}

\subsection{Ablation Studies}

We conducted a series of ablation studies to investigate the impact of different training configurations on the model's performance. Table \ref{tab:ablations} summarizes the results of these experiments.

\begin{table}[ht]
    \centering
    \caption{Ablations on training model with/without introducing new transcribe token}
    \label{tab:ablations}
    \begin{tabular}{l S[table-format=1.0] S[table-format=1.0] S[table-format=1.0] S[table-format=1.0] S[table-format=1.3]}
    \toprule
    \textbf{Test Name} & {\textbf{Transcribe token}} & {\textbf{SpeechQA}} & {\textbf{Instruction}} & {\textbf{Transcription}} & {\textbf{MMLU}} \\
    \midrule
    Recovery test 1 & 1 & 1 & 1 & 0 & 0.515 \\
    Recovery test 2 & 1 & 1 & 1 & 1 & 0.480 \\
    Recovery test 3 & 0 & 1 & 1 & 1 & \textbf{0.630} \\
    \bottomrule
    \end{tabular}
\end{table}

The results from our ablation studies provide interesting insights into the role of transcription tokens and data in model performance. Notably, Test 3, which used transcription prompts without a specific transcription token, achieved the highest MMLU score of 0.63. This suggests that the inclusion of diverse transcription prompts in the training data may be more beneficial than using a dedicated transcription token.

Interestingly, Test 1, which excluded transcription data entirely, outperformed Test 2, which included both transcription tokens and data. This unexpected result warrants further investigation and may indicate potential interactions between different types of training data that affect model performance.

These findings highlight the complex relationships between training data composition, token usage, and model performance. Future work could explore these relationships in more detail, potentially leading to improved strategies for training multi-modal language models.

\end{document}

%% file: 1-introduction.tex
\section{Introduction}\label{intro}

Large Language Models (LLMs) have become powerful tools for solving general tasks, helping people in daily life through conversations \citep{openai2024gpt4technicalreport, brown2020language, hoffmann2022training, touvron2023llama, radford2019language}. While these models have transformed text-based interactions, audio remains essential for human communication, carrying information that often exceeds written text.

Most voice assistants use a cascaded system architecture. In this approach, a user triggers an automatic speech recognition (ASR) system for transcribing the request to text. Then, A natural language understanding (NLU) pipeline converts this query into a structured format, used to generate a text answer through natural language generation (NLG). Finally, a text-to-speech (TTS) system vocalizes the answer to the user. This process, with its multiple steps, often leads to high latency that reduce user experience.

Despite improvements, these interfaces still fall short of natural conversations. First, The multiple steps in these systems add up to several seconds of delay. This contrasts with natural conversations, where responses typically come within milliseconds. Second, complexity deployment in edge device (model compatible with conventional method and not cascaded system).

Recent models that handle multiple types of data have become popular, but they still process different data types separately. This can limit how well they combine information from different sources and create documents that mix speech and text.

In this paper, we present Ichigo, a mixed-modal model capable of generating and reasoning with mixed sequences of arbitrarily interleaved textual and speech content. This approach allows for complete modeling of documents with multiple data types, expanding on tasks like understanding speech and text-only language models.

While similar ideas have been tested with images, they haven't been fully explored with audio \citep{team2024chameleon}. Previous research has trained entire models from scratch, including the LLM. Although this method is optimal, it is expensive and challenging for many research labs to adapt and build upon. Our approach, in contrast, utilizes current strong open-source LLMs and extends their capability to speech through continual pre-training. This solution not only achieves the goal of introducing a new modality to the model but also offers greater flexibility for adaptation with other LLMs family in the field.

Ichigo is designed to be mixed-modal from the start, employing a uniform architecture in an end-to-end fashion on an interleaved mixture of modalities: speech and text. By quantizing speech into discrete tokens, allowing us to use a decoder-only transformer architecture for both speech and text tokens, without adding a speech encoder and a speech adaptor \citep{fang2024llama,chu2024qwen2, tang2023salmonn, ramesh2022hierarchical}. This approach projected different data types into a shared representational space from the start, allows for smooth reasoning and generation across modalities. It represents a significant advancement over traditional cascaded systems and even recent multimodal models that treat modalities separately.

We summarize our contributions as follows:

1. We present Ichigo, an tokenized early-fusion multimodal model capable of reasoning over and generating interleaved speech-text documents.

2. We introduce training techniques for tokenized early-fusion multimodal models without starting from scratch, making our approach more accessible and adaptable.  

3. We present a recovering capability training method and techniques to stabilize cross-modality training, enhancing the robustness of our model.

4. We construct and release Instruction Speech \citep{InstructionSpeech2024}, a large-scale English speech-text cross-modal instruction-following dataset featuring multi-turn interactions, reasoning tasks, and refusal scenarios. We also provide the training and inference code to facilitate further research in this area.

%% file: 2-model-arch.tex
\section{Model Architecture}\label{model-arch}

\subsection{Tokenized Early Fusion}

\begin{figure}[ht]
\centering
\includegraphics[width=\textwidth]{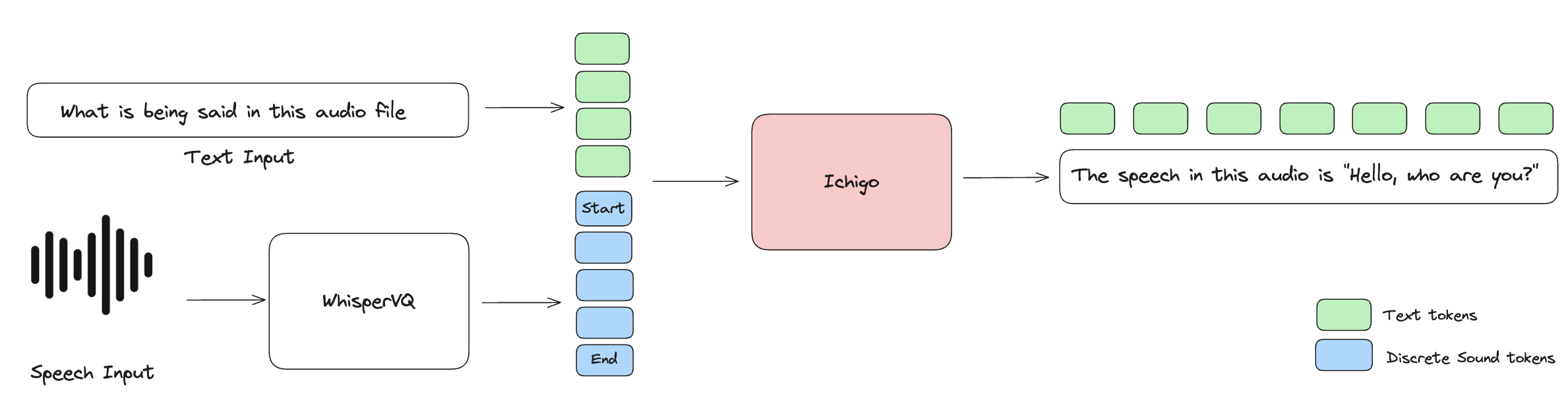}
\caption{Ichigo represents speech and text modalities as discrete tokens and uses a uniform transformer-based architecture. It uses WhisperVQ to quantize speech into discrete tokens in the same manner with original text modality.}
\label{fig:ichigo-architecture}
\end{figure}

This methodology presents a unified framework leveraging token-based representations for both speech and textual modalities (Figure \ref{fig:ichigo-architecture}). By quantizing continuous speech into discrete tokens, similar to words in text, we can utilize the same transformer architecture to sequences of both speech and text tokens. This eliminates the need for separate speech/text encoders or domain-specific decoders. By projecting all modalities into a shared representational space from the outset, this method facilitates cross-modal reasoning and generation.

\subsection{Tokenization Process}

For speech tokenization, we employ WhisperVQ, a component of WhisperSpeech \citep{WhisperSpeech2024}. This model utilizes a codebook of 512 tokens with a codebook dimension of 64. Based on the Whisper Medium model, WhisperVQ processes speech input resampled to 16 kHz, achieving a frame rate of 25 Hz.

Initially, the audio is converted to a log-mel spectrogram and processed by a Whisper encoder \citep{Whisper}, producing continuous embeddings. These embeddings undergo downsampling and refinement before a vector quantization step maps them to a finite codebook, producing a sequence of discrete tokens representing the audio content.

\subsection{Expanding the Language Model}

To incorporate multimodal discrete representations into pre-trained LLMs, we expand the vocabulary with new modality-specific tokens. This expansion necessitates extending the corresponding embeddings and prediction layer, with newly incorporated parameters initialized randomly. The combined tokens from all modalities form a new vocabulary, where each modality is trained within the language model to align in a shared representational space.

This approach allows us to compress multimodal data into discrete token sequences, which the language model can then train using next token prediction loss. Consequently, this enables the LLM to unify tasks such as understanding, reasoning, and generation in an autoregressive manner.

\subsection{Model Implementation Details}

We use Llama-3.1-8B-Instruct as our backbone model, which has been pre-trained on 15 trillion text tokens and performs well across benchmarks \citep{dubey2024llama}. Apart from reshaping the embedding matrix to accommodate the new tokens, the rest of the language model remains unaltered. The tokens generated by WhisperVQ are converted to the format <|sound\_dddd|>, where 'dddd' represents the position of the corresponding code. Additionally, we introduce two new special tokens, <|sound\_start|> and <|sound\_end|>, to delimit audio file inputs.

Initially, we attempted to use the default new token initialization from the HuggingFace codebase. However, this approach resulted in slow convergence of the loss curve. To address this issue, we switched to initializing new token embeddings by averaging all embeddings of the current vocabulary \citep{hewitt2021initializing}. This modification significantly improved the speed of convergence and enhanced training stability.

%% file: 3-data.tex
\section{Datasets}\label{data}

To enable Ichigo to process and understand audio signals, we have curated a comprehensive dataset comprising two main components: the Pre-training Dataset and the Instruction Speech Dataset. The former facilitates the LLM's understanding of audio signals, while the latter enables cross-modal instruction tuning. This section provides a detailed overview of our data collection and processing methodologies.

\subsection{Pre-training Dataset}

To align the embeddings of text and audio, we assembled a diverse collection of public Automatic Speech Recognition (ASR) datasets spanning eight languages: English, German, Dutch, Spanish, French, Italian, Portuguese, and Polish. We obtained English from the MLS English 10k dataset \citep{Pratap2020MLSAL} and other languages from the Multilingual LibriSpeech dataset \citep{Pratap2020MLSAL}.

The training dataset encompasses approximately 10,000 hours of English audio and an additional 6,000 hours distributed across the other languages. The majority of this audio content originates from audiobooks available on LibriVox and OpenSLR \citep{panayotov2015librispeech}.
Subsequently, we employed WhisperVQ to convert these audio files into discrete sound tokens.

\subsection{Post-training Dataset}

\subsubsection{Text Instruction Data}

Our training dataset comprises a blend of high-quality, open-source data available on HuggingFace \citep{xu2024magpie, huggingfacetb2024everyday, pjmixers2024math, euclaise2024gsm8k, intel2024orca, routellm2024gpt4, nomicai2024gpt4all, microsoft2024orca, allenai2024wildchat, openorca2024oogpt4, magpiealign2024magpie, qiaojin2024pubmedqa, undi952024capybara, hannahrosekirk2024prism, baai2024infinity}. These datasets span a wide array of topics, thereby diversifying the input data for our model. We implemented a two-step filtering process to ensure data quality and relevance. The main steps are Language Identification and Deduplication.

\textbf{Language Identification:} We applied the FastText model \citep{bojanowski2017enriching} as a language identifier at the document level, retaining only English documents with a confidence threshold of (0.9). This decision aligns the model's distribution more closely with the original multilingual training of the base LLM.

\textbf{Deduplication:} We removed duplicate entries to prevent overfitting and ensure a diverse training set. Despite the tokenizer's capacity to handle eight languages, we opted to focus primarily on English for this training iteration. This decision was motivated by the relative scarcity of high-quality instruction data in other languages and the low-resource nature of these languages in our dataset.

\subsubsection{Speech-Text Instruction Data}

\begin{figure}[ht]
\centering
\includegraphics[width=0.9\textwidth]{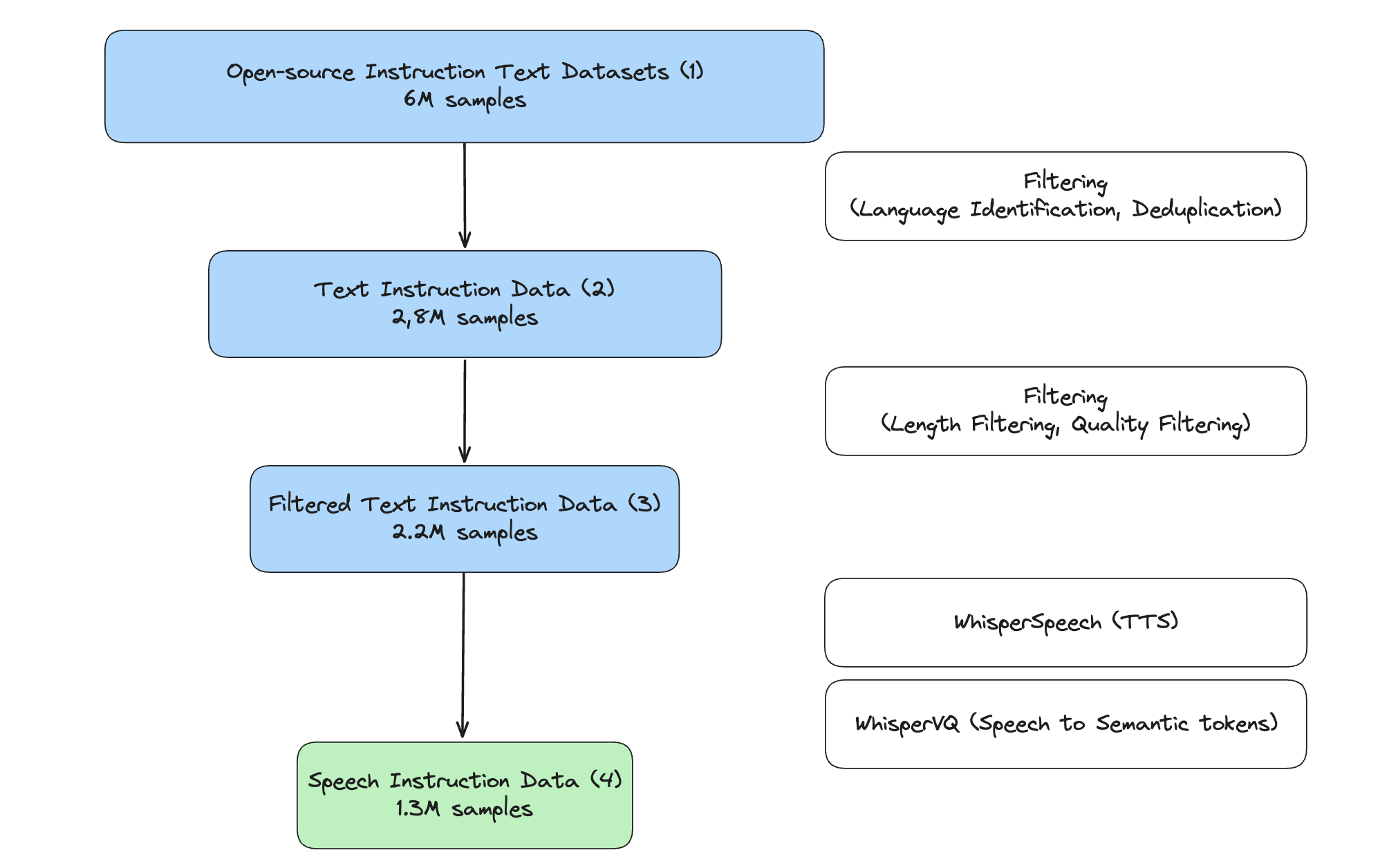}
\caption{Data Processing Pipeline for Speech Instruction Dataset Generation. This chart illustrates the multi-stage filtering and conversion process, starting from 6M samples of multiple open-source instruction text datasets. The data undergoes filtering process results in 2.2M samples. Finally, these samples are converted to speech instruction data using WhisperSpeech (TTS) and WhisperVQ (speech to semantic tokens), creating the 1.3M pairs of Speech instruction and Text answer.}
\label{fig:data-processing-pipeline}
\end{figure}

Building upon the Text Instruction Dataset, we conducted further filtering to create a dataset more suitable for Instruction Speech Dataset (Figure \ref{fig:data-processing-pipeline}).

\textbf{Length Filtering:} To prevent exceeding the LLM's context length, we filtered out text instructions longer than 64 tokens. This threshold was established based on empirical observations of typical user interactions with audio assistants.

\textbf{Quality Filtering:} We eliminated samples that would be challenging to pronounce as speech, such as URLs, mathematical symbols, and code snippets.

\textbf{Synthetic Data Generation Pipeline:} We implemented a two-stage process to convert our text-based Instruction dataset into discrete sound tokens suitable for audio input. We utilized the WhisperSpeech text-to-speech (TTS) model to generate audio files from the instruction dataset's questions. Subsequently, we employed the WhisperVQ model to transform these audio files into discrete sound tokens. Figure \ref{fig:data-processing-pipeline} illustrates the overview of the synthetic data generation pipeline.

This process was applied only to the input questions, while the corresponding answers were maintained in their original text format. The resulting dataset comprised 2000 hours of tokenized speech audio data paired with text responses.

This approach allowed us to create a rich, multimodal dataset that closely mimics real-world interactions with audio-based AI assistants, enhancing our model's ability to process and respond to spoken instructions.

\subsubsection{Transcribe Data}

For transcription tasks, we created a specialized transcribe instruction dataset derived from our ASR dataset. We introduced a signal to help the model identify transcription tasks. Initially, we experimented with a special token <|transcribe|>, but this approach led to catastrophic forgetting in the model (Table \ref{tab:ablations}).

To address this issue, we transitioned to using pure instructions. We incorporated six instruction sentences for transcription tasks, which improved the model's ability to map sound token patterns to corresponding text while minimizing the reduction in the model's text capabilities. Examples of these instructions are provided in Table \ref{tab:asr-prompts}.

\subsubsection{Noise Audio Data}

During model training, we recognized the need for noise data to prevent the model from being overly sensitive to inaudible inputs. Our initial approach of creating a synthetic dataset of random environmental noises proved challenging to scale.

We hypothesized that meaningful speech follows certain patterns and utilized this insight to generate inaudible input data. Using the 512 sound tokens from the WhisperVQ codebook, we randomized them into patterned sequences. This method allowed us to generate a vast amount of inaudible input data with a wide distribution. We then employed the Qwen2.5-72B model \citep{qwen2.5} to generate diverse synthetic answers for those inaudible inputs.

With an average speech input of about 50 sound tokens, there are $513^{50}$ possible arrangements, of which only a tiny fraction would constitute meaningful speech. By exposing our model to a wide range of these chaotic arrangements, we taught it to distinguish between audible and inaudible inputs effectively.

We also performed sequence length distribution matching between inaudible and audible data to ensure a balanced representation of both types of inputs in our training set. This approach involved sampling inaudible data samples to match the token count distribution of the original data, contributing to a more robust and generalizable model.

%% file: 4-training.tex
\section{Training}

The training process for our model was conducted in multiple stages, each designed to optimize different aspects of performance and functionality. This section details the software infrastructure, pre-training methodology, and post-training refinements employed in our research.

\subsection{Pre-training Methodology}

Our pre-training approach was rooted in the fundamental principle of language model training: converting text into processable tokens and enabling the model to learn patterns and relationships within the data. In this phase, we aimed to introduce speech representation into new tokens, facilitating the model's development of basic concepts regarding these additional tokens.

We utilized AdamW Fused optimize \citep{adamw, paszke2019pytorch} with a weight decay of 0.01, momentum decay of 0.9, and squared gradient decay of 0.95. Although we experimented with alternative optimizers such as Adam-mini and Lion during hyperparameters tuning, these attempts resulted in unstable training and frequent loss explosions, prompting our return to AdamW Fused.

All models were trained on our internal cluster comprising 10 NVIDIA A6000-48GB GPUs, employing FSDP 2 \citep{torchtune} and activation checkpointing. The training consisted of 8,064 steps with a batch size of 480 and a context length of 512. We implemented a Cosine learning rate schedule \citep{torchtune} initiating at 2e-4 with warmup over 50 steps.

Table \ref{tab:training-hyperparameters} provides an overview of the hyper-parameters and configurations used in the three-stage training phases. In the Pre-training stage, we maximize the global batch size to ensure more general learning of the model. During Instruction Fine-tuning and Enhancement Fine-tuning, we reduce the learning rate to stabilize the training loss curve. Additionally, we increase the context length to 4096 tokens in these later stages, providing the model with more space to respond to user requests.

\begin{table}[ht]
    \centering
    \caption{Training Hyper-parameters for Ichigo's three-stage process.}
    \label{tab:training-hyperparameters}
    \begin{tabular}{lccc} \toprule
        \textbf{Parameter} & \textbf{Pre-training} & \textbf{Instruction FT} & \textbf{Enhancement FT} \\ \midrule
        Weight Decay & \multicolumn{3}{c}{0.005} \\
        Learning Scheduler & \multicolumn{3}{c}{Cosine} \\
        Optimizer & \multicolumn{3}{c}{AdamW Fused} \\
        Precision & \multicolumn{3}{c}{bf16} \\ \addlinespace
        Hardware & 10x A6000 & 8x H100 & 8x H100 \\
        Train time & 45h & 10h & 3h \\
        Steps & 8064 & 7400 & 644 \\
        Global batch size & 480 & 256 & 256 \\
        Learning Rate & $2 \times 10^{-4}$ & $7 \times 10^{-5}$ & $1.5 \times 10^{-5}$ \\
        Warmup Steps & 50 & 73 & 8 \\
        Max length & 512 & 4096 & 4096 \\ \bottomrule
    \end{tabular}
\end{table}

\subsection{Post-training Refinements}

The post-training phase was divided into two distinct stages: instruction fine-tuning and enhancement fine-tuning. The former focused on honing the model's question-answering capabilities, while the latter expanded its proficiency in multi-turn conversations and appropriate responses to inaudible inputs.

\subsubsection{Instruction Fine-tuning}

Building upon the model from the previous stage, we concentrated on developing its question-answering abilities. Our research revealed the critical importance of balancing modalities during the Supervised Fine-Tuning (SFT) stage to maintain the model's original performance. We observed that significant imbalances between modality pairings could lead to unconditional priors, resulting in either muted or exaggerated generation of specific modalities.

\begin{figure}[ht]
\centering
\includegraphics[width=1.0\textwidth]{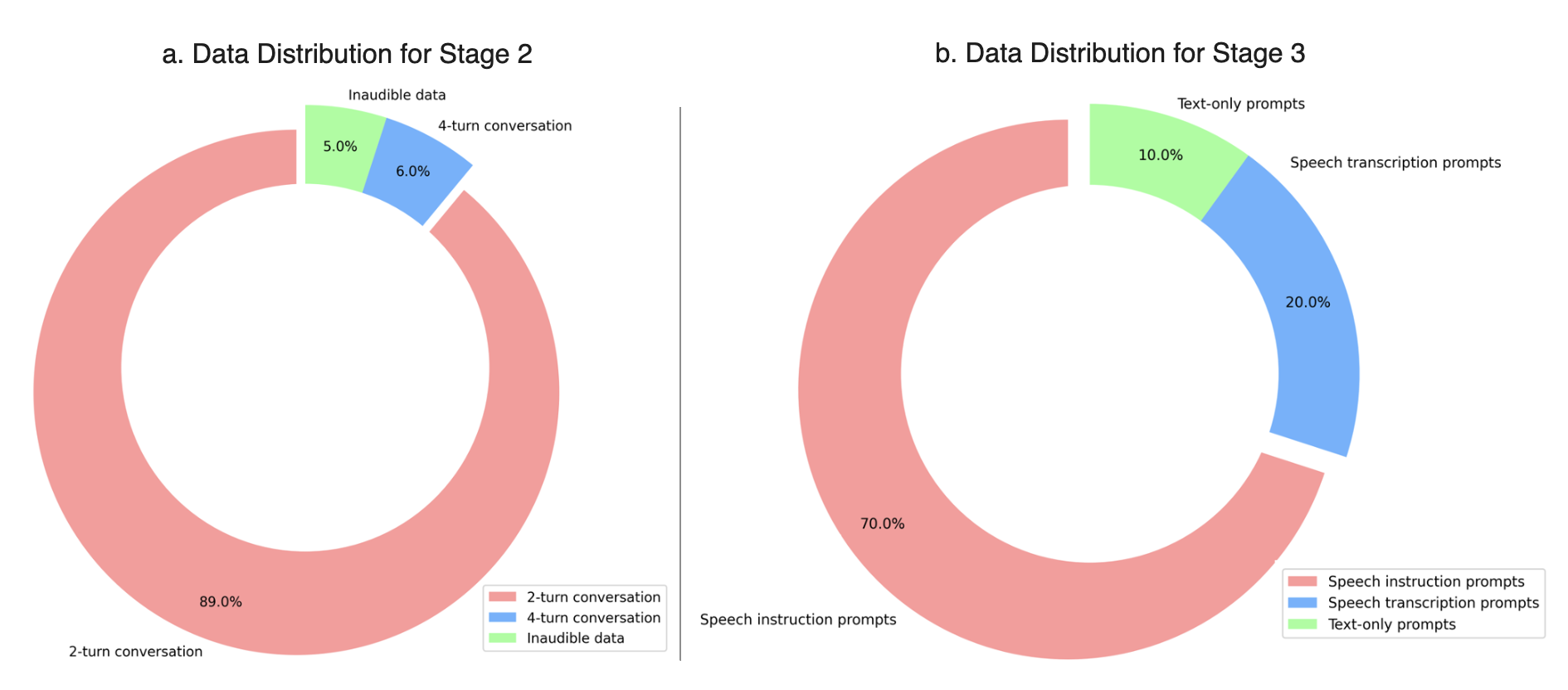}
\caption{\textbf{a.} Distribution of data types in the Instruction Fine-tuning dataset. The goal of this specific distribution was to enhance speech comprehension while maintaining robust general language abilities. \textbf{b.} Distribution of data samples used in the enhancement fine-tuning stage. This specific distribution improves Ichigo robustness in handling multi-turn conversations and inaudible inputs.}
\label{fig:instruction-finetuning}
\end{figure}

To address this, we carefully curated our dataset, comprising 70\% speech instruction prompts, 20\% speech transcription prompts, and 10\% text-only prompts. This distribution was determined through extensive permutation testing to achieve an optimal balance between speech understanding, transcription capabilities, and general language skills. Figure \ref{fig:instruction-finetuning} shows the data distribution proportion for this training stage.

\subsubsection{Enhancement Fine-tuning}

The enhancement fine-tuning stage involved data augmentation to simulate real-world user interactions, thereby improving Ichigo's robustness in various scenarios. We focused on two key areas: multi-turn conversations with speech input and appropriate responses to inaudible inputs. These enhancements aimed to create more fluid dialogues and improve the model's interactive capabilities.

To achieve this, we fine-tuned the model using a dataset of 158,000 samples. The dataset for enhancing refusal capabilities was carefully balanced, comprising only 0.5\% of the total multi-turn data. This proportion was determined through experimentation, as we found that a higher percentage led to an increased tendency for the model to refuse inputs. Figure \ref{fig:instruction-finetuning} illustrates the data distribution ratios.

%% file: 5-results.tex
\section{Results}

In this section, we present the experimental outcomes for Ichigo. We evaluate its performance across multiple dimensions, including question-answering capabilities, response latency, degradation recovery, and practical cases. Our analysis provides a comprehensive assessment of Ichigo's capabilities in comparison to other well-known speech language models.

\subsection{SpeechBench Evaluation}

We first assess Ichigo's speech question-answering (SQA) ability in comparison to other well-known speech language models. Table \ref{tab:sqa-comparison} presents the results on two SQA scores from AudioBench \citep{wang2024audiobench}, using the robust LLaMA-3 70B model \citep{dubey2024llama} as the judge for evaluation.

\begin{table}[ht]
    \centering
    \caption{A comparative results of Ichigo against three representative Speech Language Models and a cascade system.}
    \label{tab:sqa-comparison}
    \begin{tabular}{lSS} \toprule
        \textbf{Model} & \textbf{OpenHermes-Audio} & \textbf{ALPACA-Audio} \\ \midrule
        Whisper + Llama-3 8B & 63.0 & \textbf{70.8} \\
        \addlinespace
        \addlinespace
        SALMONN & 19.2 & 12.4 \\
        Qwen2-Audio & 44.8 & 52.0 \\
        WavLM & 22.4 & 21.6 \\
        Ichigo instruct v0.3 (Phase 3) & \textbf{67.8} & 67.2 \\ \bottomrule
    \addlinespace[0.5em]
    \multicolumn{3}{l}{\textit{\textsuperscript{*}Note: Higher scores indicate better performance.}} \\
\end{tabular}
\end{table}

It is important to note that during our evaluation, we encountered an error in the judge model's output affecting the 'Rating score'. The model provided ratings in the middle of its responses rather than at the end as expected, resulting in lowered scores. To address this issue, we implemented a backfilling procedure for missing ratings, ensuring a more accurate representation of model performance.

Our results demonstrate that Ichigo outperforms existing open-source speech language models, particularly those utilizing Non-Tokenized Early Fusion (NTEF) approaches \citep{wadekar2024evolution}. Compared to other end-to-end models, Ichigo's performance is particularly impressive. It outperforms Qwen2-Audio \citep{chu2024qwen2}, the next best performer among end-to-end models, by 23 points on OpenHermes-Audio and 15.2 points on ALPACA-Audio. This substantial improvement underscores the effectiveness of Ichigo's architecture and training approach in capturing the nuances of speech-language interactions.

On the OpenHermes-Audio benchmark, Ichigo achieves a score of 67.8, surpassing even the cascaded system (63.0). This performance is especially noteworthy given that cascaded systems often benefit from specialized components for transcription and language modeling.

For the ALPACA-Audio benchmark, Ichigo maintains its strong performance with a score of 67.2. While this is lower than the cascaded system (70.8), it's important to note that Ichigo achieves this as an end-to-end model, without the need for separate transcription and language modeling phases. This demonstrates Ichigo's ability to effectively integrate speech understanding and language generation in a single model.

\subsection{Latency to first token}

To validate the efficiency of Ichigo's Tokenized Early Fusion architecture, we conducted a comparative analysis of its latency to first token against current speech models and cascaded systems. Our benchmarking was performed on a single NVIDIA A6000-48GB GPU, performing 10 iterations of the latency test. The test set comprised 10 diverse audio files with durations ranging from 1 to 5 seconds (average length: 5.4 $\pm$ 2.79 seconds), reflecting real-world usage scenarios. This setup ensures a comprehensive evaluation across various audio lengths.

\begin{table}[ht]
    \centering
    \caption{The comparative results of latency to first token and VRAM usage across different models and systems}
    \label{tab:latency-comparison}
    \begin{tabular}{l S[table-format=3.2] @{${}\pm{}$} S[table-format=2.5] S[table-format=2.0]}
    \toprule
    \textbf{Model} & \multicolumn{2}{c}{\textbf{Latency (avg.)}} & {\textbf{VRAM usage}} \\
    & \multicolumn{2}{c}{(ms)} & {(GB)} \\
    \midrule
    Qwen2-Audio & 317.45 & 8.30 & 32 \\
    Cascaded system & 453.18 & 15.02 & 19 \\
    Ichigo & \bfseries \textbf{111.52} & \bfseries 7.73 & \bfseries \textbf{19} \\
    \bottomrule
    \end{tabular}
    \end{table}

\textbf{Efficiency of Direct Generation:} Our pipeline, which generates responses directly from the model, significantly reduces the latency to first response compared to the cascaded system. Ichigo achieves an average latency of 111.52 $\pm$ 7.73 ms, which is approximately 4 times faster than the cascaded Whisper + Llama-3 8B system (453.18 $\pm$ 15.02 ms).

\textbf{Comparison with Other Speech Language Models:} Benefiting from LLM-specific inference engines, Ichigo outperforms other speech language models in terms of latency. Ichigo achieves a 110 ms latency to first response, which is nearly 3 times faster than Qwen2-Audio (317.45 $\pm$ 8.30 ms).

\textbf{VRAM Efficiency:} Ichigo maintains a lower VRAM footprint (19 GB) compared to Qwen2-Audio (32 GB) and the cascaded system (19 GB). This demonstrates Ichigo's exceptional efficiency in balancing high performance with resource utilization.

\subsection{Degradation recovery}

In the process of training multi-modal models, a critical concern is not only how well the model learns new modalities but also how effectively it retains the capabilities of its original language model. To assess this, we evaluated Ichigo on three popular LLM benchmarks spanning a wide range of topics including General Knowledge, Reasoning, and Mathematics, using the LM Evaluation Harness \citep{eval-harness}.

Table \ref{tab:degradation-recovery} presents the comparative results of Ichigo across different versions and the original Llama3 8B Instruct model. The metrics used are MMLU (5-shot) \citep{hendrycks2020measuring}, GPQA (0-shot) \citep{rein2023gpqa}, and GSM-8K (Chain-of-Thought, 8-shot) \citep{cobbe2021training}, which provide a comprehensive evaluation of the model's capabilities across various domains.

\begin{table}[ht]
    \centering
    \caption{Results of Ichigo across different versions and the original Llama3 8B Instruct model.}
    \label{tab:degradation-recovery}
    \begin{tabular}{l S[table-format=2.2] S[table-format=2.2] S[table-format=2.2] S[table-format=2.2]}
    \toprule
    \textbf{Model} & {\textbf{MMLU}} & {\textbf{GPQA}} & {\textbf{GSM-8K}} & {\textbf{Avg.}} \\
    & {\textbf{(5-shots)}} & {\textbf{(0-shot)}} & {\textbf{(CoT) (8-shots)}} & \\
    \midrule
    \addlinespace
    Llama3 8B Instruct & 69.4 & 30.4 & 84.5 & 61.43 \\
    \addlinespace
    \addlinespace
    Ichigo base v0.2 & 47.66 & 28.13 & {N/A\textsuperscript{*}} & {N/A\textsuperscript{*}} \\
    Ichigo instruct v0.2 & 50.27 & 26.56 & 53.58 & 43.47 \\
    Ichigo base v0.3 & 42.11 & 28.57 & {N/A\textsuperscript{*}} & {N/A\textsuperscript{*}} \\
    {Ichigo instruct v0.3} & 63.08 & 28.35 & 76.50 & 55.98 \\
    \multicolumn{1}{l}{(phase 2)} & & & & \\
    Ichigo instruct v0.3 & \bfseries 63.79 & \bfseries 29.69 & \bfseries 75.28 & \bfseries 56.25 \\
    \multicolumn{1}{l}{(phase 3)} & & & & \\
    \bottomrule
    \addlinespace[0.5em]
    \multicolumn{5}{l}{\textit{\textsuperscript{*}N/A: Not applicable due to significant performance degradation in}} \\
    \multicolumn{5}{l}{\textit{mathematical and coding tasks during pre-training on sound tokens with}} \\
    \multicolumn{5}{l}{\textit{next token prediction.}} \\
    \end{tabular}
\end{table}

Our findings reveal a significant improvement in performance retention from earlier versions to the latest Ichigo instruct v0.3 (phase 3). Notably, the final training phase of Ichigo achieved a reduction in performance degradation from 29.3\% (in v0.2) to only 8.4\% (in v0.3 phase 3) compared to the original Llama3 8B Instruct model. This substantial recovery is primarily attributed to our refined training strategy, which incorporates a mixed proportion of text-only and sound token data.

\textbf{Performance Recovery:} Ichigo instruct v0.3 (phase 3) demonstrates remarkable recovery across all benchmarks. For instance, on the MMLU benchmark, it achieves a score of 63.79, significantly closer to the original Llama3 8B Instruct's 69.4, compared to the 50.27 scored by v0.2.

\textbf{Consistent Improvement:} We observe a consistent upward trend in performance from v0.2 to v0.3 (phase 3), indicating the effectiveness of our iterative training approach. This indicates that with extended training time and more computational resources, we can achieve higher performance.

\textbf{Pre-training Challenges:} It's worth noting that during the initial pre-training phase, which focused solely on sound tokens with next token prediction, we observed a significant degradation in the model's performance on text-based tasks, particularly in mathematics and coding. This highlights the challenges in maintaining cross-modal capabilities during specialized training.

\subsection{Instruction following cross modality}

In addition to the quantitative results presented earlier, we conducted practical experiments with Ichigo in real-world conversational scenarios. These experiments aimed to assess the model's ability to follow system prompts and maintain coherent multi-turn dialogues across different modalities (text and speech).

\begin{figure}[ht]
\centering
\includegraphics[width=0.6\textwidth]{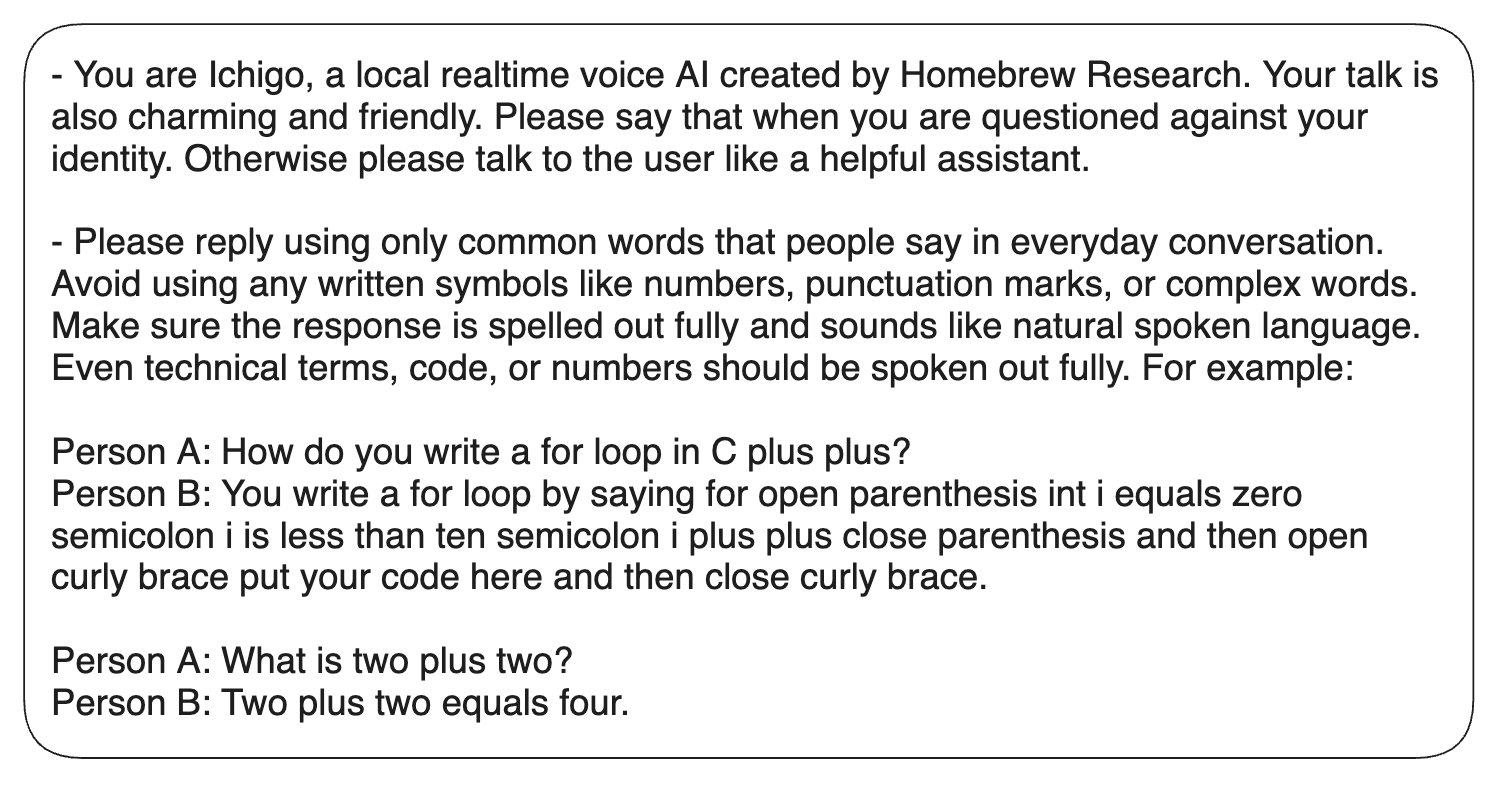}
\caption{The system prompt used for Ichigo during inference.}
\label{fig:system-prompt}
\end{figure}

\textbf{Cross-Modal Instruction Following:} Ichigo demonstrated a robust ability to follow text-based system prompts while engaging in speech-based conversations with users. This highlights the model's capacity to generalize instructions across modalities, a crucial feature for versatile AI assistants.

\begin{figure}[ht]
    \centering
    \includegraphics[width=0.8\textwidth]{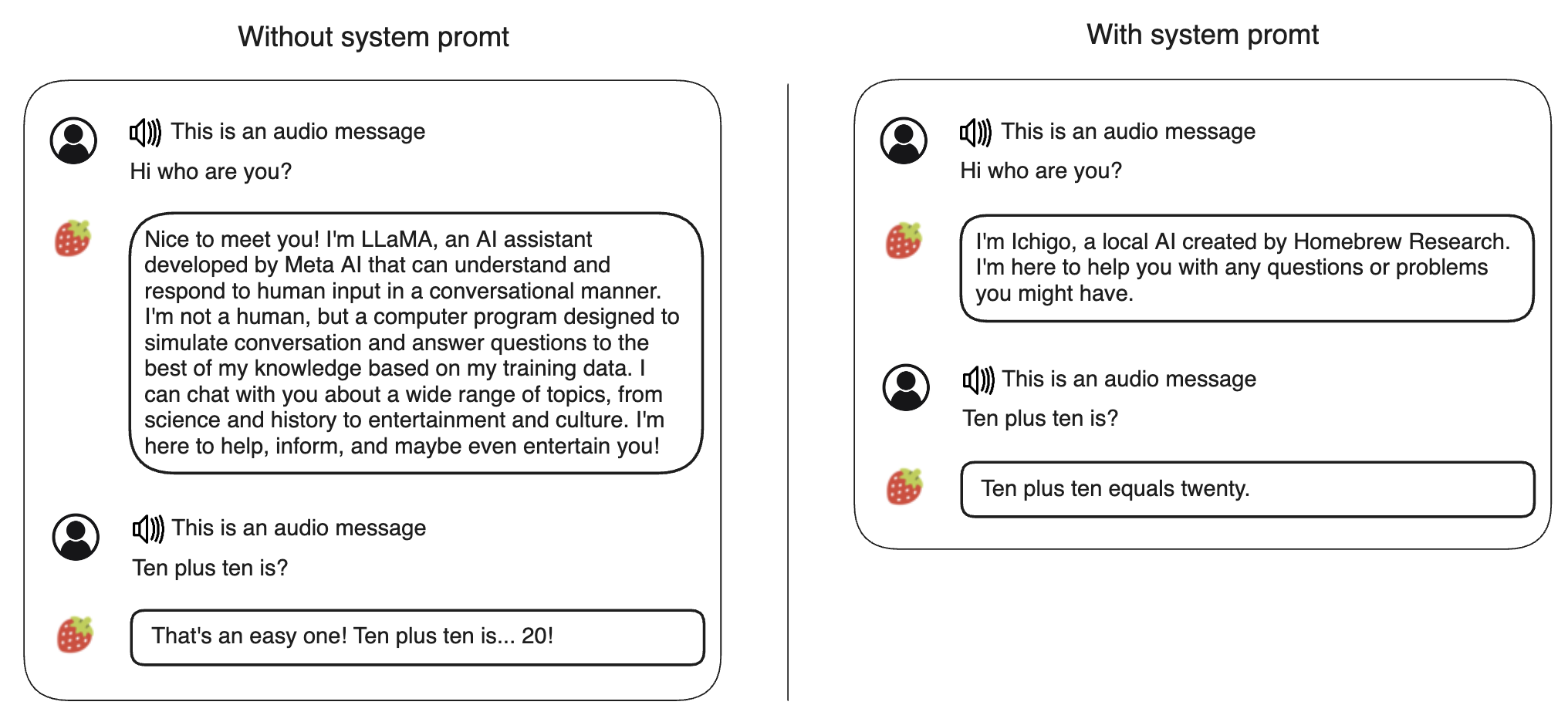}
    \caption{The model follows text-based system prompts during speech-based conversations with users.}
    \label{fig:cross-modal-instruction}
    \end{figure}    

As shown in Figure \ref{fig:cross-modal-instruction}, the model consistently maintained its prescribed identity as "Ichigo" when questioned, adhering to the system prompt instructions. This behavior persisted regardless of whether the input was in text or speech format, demonstrating the model's ability to maintain context across different input modalities.

\begin{figure}[ht]
\centering
\includegraphics[width=0.8\textwidth]{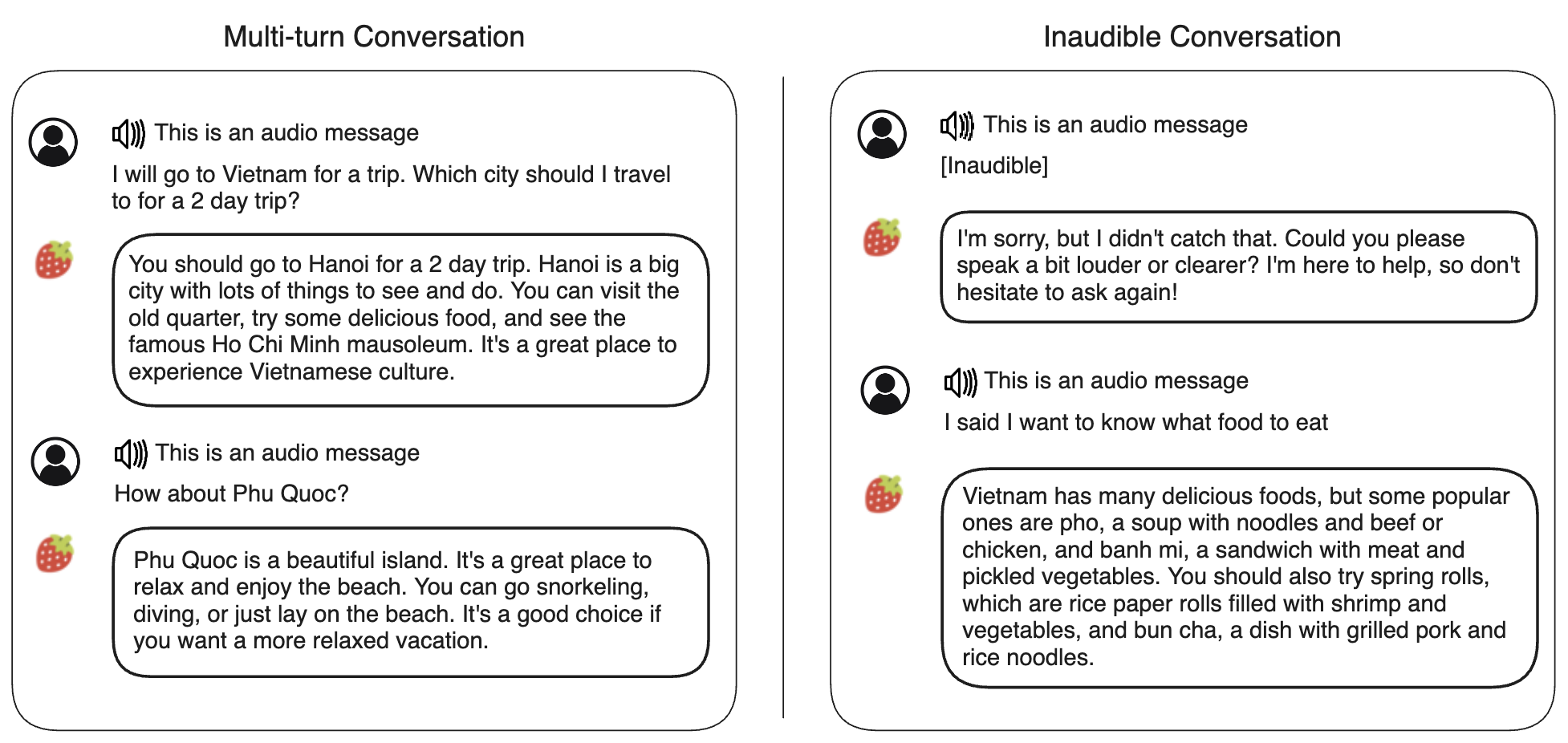}
\caption{\textbf{a.} Transcribed dialogue examples using Ichigo. The user-turn is audio input. The examples illustrate zero-shot multi-turn capabilities. \textbf{b.} Ichigo requests clarification from the user when unable to understand the question clearly.}
\label{fig:multi-turn-dialogue}
\end{figure}

\textbf{Multi-Turn Dialogue Coherence:} Ichigo exhibited proficiency in managing multi-turn conversations, seamlessly understanding and responding to both speech and text inputs without apparent difficulties. Figure \ref{fig:multi-turn-dialogue} presents transcribed dialogue examples that showcase the model's zero-shot multi-turn capabilities.

\textbf{Handling Unclear Inputs:} In scenarios where user input was inaudible, unclear, or affected by background noise, Ichigo demonstrated appropriate behavior by refusing to provide random answers. Instead, as illustrated in Figure \ref{fig:multi-turn-dialogue}, the model politely requested the user to repeat their query, ensuring accurate and relevant responses.

These experiments complement our quantitative findings, demonstrating Ichigo's practical capabilities in real-world scenarios. Ichigo's abilities in cross-modal instruction following, multi-turn dialogues, and handling unclear inputs make it a promising candidate for advanced, user-friendly voice AI applications.

%% file: 6-conclusion.tex
\section{Related works}

\subsection{Early Audio Language Models}

The success of language models in natural language processing \citep{radford2019language, raffel2020exploring, openai2024gpt4technicalreport} has inspired researchers to explore similar approaches for modeling speech and audio. Initial efforts in audio language modeling focused on training models using semantic or acoustic tokens derived from audio data, enabling audio generation without the need for text input \citep{borsos2023audiolm, nguyen2023generative, lakhotia2021generative}. Subsequent advancements led to the joint training of speech tokens and text, resulting in decoder-only models such as VALL-E \citep{wang2023neural, chen2024vall} and VioLA \citep{wang2023viola}. These models demonstrated capabilities in speech recognition, translation, and synthesis. However, these early models were not built upon Large Language Models (LLMs). To harness the power of LLMs, researchers have explored various approaches to building speech-language models based on LLM architectures.

\subsection{LLM-Based Audio-Language Models}

Recent research has focused on two primary approaches to integrating speech and audio capabilities with LLMs: non-tokenized early fusion and tokenized early fusion.

\subsubsection{Non-Tokenized Early Fusion}

The most common approach to enable cross-modal perception in LLMs is to connect pre-trained encoders of other modalities as adaptors. This method involves adding a speech encoder before the LLM and fine-tuning the entire model for speech understanding capabilities. These models excel in tasks such as speech recognition, speech translation, and general speech-to-text tasks \citep{chu2024qwen2, tang2023salmonn, shu2023llasm, deshmukh2023pengi, hu2024wavllm, das2024speechverse, fang2024llama}. Notable examples of this approach are Llama-omni \citep{fang2024llama} and LLaSM \citep{shu2023llasm}, which extend LLM capabilities to audio modality by integrating a pre-trained speech encoder, a speech adaptor, and a streaming speech decoder. SALMONN \citep{tang2023salmonn} takes a step further in capturing both speech and non-speech audio information using dual auditory encoders. Qwen2 Audio \citep{chu2024qwen2} introduce the new architecture to combine an audio encoder with a large language model, training to maximize next text token probability conditioned on audio representations.

This NTEF tends to be more cost-effective, as it involves multiple training phases where most components are frozen, and it can be effective even when training with Parameter-Efficient Fine-Tuning techniques \citep{hu2021lora}.

\subsubsection{Tokenized Early Fusion}

This approach involves tokenizing multimodal inputs using either a common tokenizer or modality-specific tokenizers. The tokenized inputs are then processed by a pre-trained LLM or an encoder-decoder transformer model, enabling multimodal output generation \citep{wadekar2024evolution}. Examples of this approach include Chameleon \citep{team2024chameleon}, which represents images and text as a series of discrete tokens within a unified transformer, trained from scratch with modified transformer architecture. AudioPALM \citep{rubenstein2023audiopalm} and VoxTLM \citep{maiti2024voxtlm}, which utilize pre-trained language models and extend their vocabularies with discrete semantic audio tokens, focus on translation speech to speech tasks. AnyGPT \citep{zhan2024anygpt} leverages LLMs to enable inherent cross-modal conversation capabilities through SpeechTokenizer \citep{zhang2023speechtokenizer}, MusicTokenizer \citep{defossez2022highfi}, and ImageTokenizer \citep{ge2023making}. Unlike previous works, our approach retains the entire architecture of current LLMs while incorporating WhisperVQ to preserve most of the OpenAI Whisper encoder block. This allows us to generate embeddings, which are then quantized to obtain semantic tokens. Additionally, we address a key challenge is to stabilize the loss in cross-modality training.

Another new approach is Moshi \citep{defossez2024moshi} is a real-time native multimodal foundation model designed for seamless audio-text interactions. It employs a 7B parameter multimodal language model that processes speech input and output concurrently, generating text tokens and audio codecs. Moshi's innovative approach allows it to handle two audio streams simultaneously, enabling it to listen and talk in real-time while maintaining a flow of textual thoughts.

\section{Conclusion}

In this paper, we introduced Ichigo, an early-fusion token-based speech model that sets a new approach for Multi-modal Models. By learning a unified representation space over interleaved speech and text tokens, Ichigo achieves strong performance across a wide range of speech-language benchmarks while enabling novel mixed-modal reasoning and generation capabilities.

The key to Ichigo's success lies in its fully token-based architecture, which allows for seamless information integration across modalities. By quantizing speech into discrete tokens and utilizing a strong base LLM, Ichigo learns to jointly reason over speech and text in a way that surpasses late-fusion architectures or models that maintain separate encoders for each modality.

Crucially, our meticulous approach to mixing training data has allowed us to largely preserve the original performance of the original LLM while extending its capabilities to the speech domain. As a result, Ichigo outperforms other end-to-end Speech Language Models in speech-based question-answering tasks, marking a significant step forward in multimodal AI.

Importantly, Ichigo demonstrates a real-time speech system with a latency of 110 milliseconds to first response. This opens up new possibilities for speech systems and significantly reduces the complexity of deploying such systems in production environments.

We believe that this paper will empower smaller research teams - like ourselves - to contribute more confidently and prolifically to the open-source community. By demonstrating that significant advancements can be achieved with limited resources, we hope to inspire broader participation in this critical area of AI research.

\section{Limitations and Future work}

While Ichigo represents a significant step forward in multimodal language modeling, several limitations and areas for future work remain:

\begin{itemize}
    \item \textbf{Token Stability}: Similar to challenges faced by models like Chameleon, we encountered fluctuating loss when training with acoustic tokens, which led us to shift towards semantic tokens to achieve stable loss. This highlights the difficulty in training with rich, acoustic information. Future work should explore methods to stabilize training with acoustic tokens, potentially unlocking even more powerful models.
    \item \textbf{Emotional Understanding}: The current architecture does not fully account for emotional comprehension. Future iterations should focus on enhancing the model's ability to understand and respond to user emotions, allowing for more nuanced and context-appropriate responses.
    \item \textbf{Context Length}: Multimodal content, especially audio, often spans extensive sequences. Ichigo currently limits modeling to 10 seconds of speech input and performs well for 4-5 turns of conversation. Extending the context window would allow for modeling of longer audio segments and handling of more complex, multi-turn conversations.
\end{itemize}

%% file: paper.bbl
\begin{thebibliography}{62}
\providecommand{\natexlab}[1]{#1}
\providecommand{\url}[1]{\texttt{#1}}
\expandafter\ifx\csname urlstyle\endcsname\relax
  \providecommand{\doi}[1]{doi: #1}\else
  \providecommand{\doi}{doi: \begingroup \urlstyle{rm}\Url}\fi

\bibitem[BAAI(2024)]{baai2024infinity}
BAAI.
\newblock Infinity-instruct, 2024.
\newblock URL \url{https://huggingface.co/datasets/BAAI/Infinity-Instruct}.

\bibitem[Bojanowski et~al.(2017)Bojanowski, Grave, Joulin, and
  Mikolov]{bojanowski2017enriching}
Piotr Bojanowski, Edouard Grave, Armand Joulin, and Tomas Mikolov.
\newblock Enriching word vectors with subword information.
\newblock \emph{Transactions of the association for computational linguistics},
  5:\penalty0 135--146, 2017.

\bibitem[Borsos et~al.(2023)Borsos, Marinier, Vincent, Kharitonov, Pietquin,
  Sharifi, Roblek, Teboul, Grangier, Tagliasacchi, et~al.]{borsos2023audiolm}
Zal{\'a}n Borsos, Rapha{\"e}l Marinier, Damien Vincent, Eugene Kharitonov,
  Olivier Pietquin, Matt Sharifi, Dominik Roblek, Olivier Teboul, David
  Grangier, Marco Tagliasacchi, et~al.
\newblock Audiolm: a language modeling approach to audio generation.
\newblock \emph{IEEE/ACM transactions on audio, speech, and language
  processing}, 31:\penalty0 2523--2533, 2023.

\bibitem[Brown(2020)]{brown2020language}
Tom~B Brown.
\newblock Language models are few-shot learners.
\newblock \emph{arXiv preprint arXiv:2005.14165}, 2020.

\bibitem[Chen et~al.(2024)Chen, Liu, Zhou, Liu, Tan, Li, Zhao, Qian, and
  Wei]{chen2024vall}
Sanyuan Chen, Shujie Liu, Long Zhou, Yanqing Liu, Xu~Tan, Jinyu Li, Sheng Zhao,
  Yao Qian, and Furu Wei.
\newblock Vall-e 2: Neural codec language models are human parity zero-shot
  text to speech synthesizers.
\newblock \emph{arXiv preprint arXiv:2406.05370}, 2024.

\bibitem[Chu et~al.(2024)Chu, Xu, Yang, Wei, Wei, Guo, Leng, Lv, He, Lin,
  et~al.]{chu2024qwen2}
Yunfei Chu, Jin Xu, Qian Yang, Haojie Wei, Xipin Wei, Zhifang Guo, Yichong
  Leng, Yuanjun Lv, Jinzheng He, Junyang Lin, et~al.
\newblock Qwen2-audio technical report.
\newblock \emph{arXiv preprint arXiv:2407.10759}, 2024.

\bibitem[Cobbe et~al.(2021)Cobbe, Kosaraju, Bavarian, Chen, Jun, Kaiser,
  Plappert, Tworek, Hilton, Nakano, et~al.]{cobbe2021training}
Karl Cobbe, Vineet Kosaraju, Mohammad Bavarian, Mark Chen, Heewoo Jun, Lukasz
  Kaiser, Matthias Plappert, Jerry Tworek, Jacob Hilton, Reiichiro Nakano,
  et~al.
\newblock Training verifiers to solve math word problems.
\newblock \emph{arXiv preprint arXiv:2110.14168}, 2021.

\bibitem[Collabora(2024)]{WhisperSpeech2024}
Collabora.
\newblock Whisperspeech, 2024.
\newblock Accessed: 19 October 2024.

\bibitem[Das et~al.(2024)Das, Dingliwal, Ronanki, Paturi, Huang, Mathur, Yuan,
  Bekal, Niu, Jayanthi, et~al.]{das2024speechverse}
Nilaksh Das, Saket Dingliwal, Srikanth Ronanki, Rohit Paturi, David Huang,
  Prashant Mathur, Jie Yuan, Dhanush Bekal, Xing Niu, Sai~Muralidhar Jayanthi,
  et~al.
\newblock Speechverse: A large-scale generalizable audio language model.
\newblock \emph{arXiv preprint arXiv:2405.08295}, 2024.

\bibitem[D{\'e}fossez et~al.(2024)D{\'e}fossez, Mazar{\'e}, Orsini, Royer,
  P{\'e}rez, J{\'e}gou, Grave, and Zeghidour]{defossez2024moshi}
Alexandre D{\'e}fossez, Laurent Mazar{\'e}, Manu Orsini, Am{\'e}lie Royer,
  Patrick P{\'e}rez, Herv{\'e} J{\'e}gou, Edouard Grave, and Neil Zeghidour.
\newblock Moshi: a speech-text foundation model for real-time dialogue.
\newblock \emph{arXiv preprint arXiv:2410.00037}, 2024.

\bibitem[Deshmukh et~al.(2023)Deshmukh, Elizalde, Singh, and
  Wang]{deshmukh2023pengi}
Soham Deshmukh, Benjamin Elizalde, Rita Singh, and Huaming Wang.
\newblock Pengi: An audio language model for audio tasks.
\newblock \emph{Advances in Neural Information Processing Systems},
  36:\penalty0 18090--18108, 2023.

\bibitem[Dubey et~al.(2024)Dubey, Jauhri, Pandey, Kadian, Al-Dahle, Letman,
  Mathur, Schelten, Yang, Fan, et~al.]{dubey2024llama}
Abhimanyu Dubey, Abhinav Jauhri, Abhinav Pandey, Abhishek Kadian, Ahmad
  Al-Dahle, Aiesha Letman, Akhil Mathur, Alan Schelten, Amy Yang, Angela Fan,
  et~al.
\newblock The llama 3 herd of models.
\newblock \emph{arXiv preprint arXiv:2407.21783}, 2024.

\bibitem[Défossez et~al.(2022)Défossez, Copet, Synnaeve, and
  Adi]{defossez2022highfi}
Alexandre Défossez, Jade Copet, Gabriel Synnaeve, and Yossi Adi.
\newblock High fidelity neural audio compression.
\newblock \emph{arXiv preprint arXiv:2210.13438}, 2022.

\bibitem[euclaise(2024)]{euclaise2024gsm8k}
euclaise.
\newblock gsm8k\_multiturn, 2024.
\newblock URL \url{https://huggingface.co/datasets/euclaise/gsm8k_multiturn}.

\bibitem[Fang et~al.(2024)Fang, Guo, Zhou, Ma, Zhang, and Feng]{fang2024llama}
Qingkai Fang, Shoutao Guo, Yan Zhou, Zhengrui Ma, Shaolei Zhang, and Yang Feng.
\newblock Llama-omni: Seamless speech interaction with large language models.
\newblock \emph{arXiv preprint arXiv:2409.06666}, 2024.

\bibitem[for AI(2024)]{allenai2024wildchat}
Allen~Institute for AI.
\newblock Wildchat-1m, 2024.
\newblock URL \url{https://huggingface.co/datasets/allenai/WildChat-1M}.

\bibitem[Gao et~al.(2024)Gao, Tow, Abbasi, Biderman, Black, DiPofi, Foster,
  Golding, Hsu, Le~Noac'h, Li, McDonell, Muennighoff, Ociepa, Phang, Reynolds,
  Schoelkopf, Skowron, Sutawika, Tang, Thite, Wang, Wang, and
  Zou]{eval-harness}
Leo Gao, Jonathan Tow, Baber Abbasi, Stella Biderman, Sid Black, Anthony
  DiPofi, Charles Foster, Laurence Golding, Jeffrey Hsu, Alain Le~Noac'h,
  Haonan Li, Kyle McDonell, Niklas Muennighoff, Chris Ociepa, Jason Phang,
  Laria Reynolds, Hailey Schoelkopf, Aviya Skowron, Lintang Sutawika, Eric
  Tang, Anish Thite, Ben Wang, Kevin Wang, and Andy Zou.
\newblock A framework for few-shot language model evaluation, 07 2024.
\newblock URL \url{https://zenodo.org/records/12608602}.

\bibitem[Ge et~al.(2023)Ge, Zhao, Zeng, Ge, Li, Wang, and Shan]{ge2023making}
Yuying Ge, Sijie Zhao, Ziyun Zeng, Yixiao Ge, Chen Li, Xintao Wang, and Ying
  Shan.
\newblock Making llama see and draw with seed tokenizer.
\newblock \emph{arXiv preprint arXiv:2310.01218}, 2023.

\bibitem[HannahRoseKirk(2024)]{hannahrosekirk2024prism}
HannahRoseKirk.
\newblock prism-alignment, 2024.
\newblock URL
  \url{https://huggingface.co/datasets/HannahRoseKirk/prism-alignment}.

\bibitem[Hendrycks et~al.(2020)Hendrycks, Burns, Basart, Zou, Mazeika, Song,
  and Steinhardt]{hendrycks2020measuring}
Dan Hendrycks, Collin Burns, Steven Basart, Andy Zou, Mantas Mazeika, Dawn
  Song, and Jacob Steinhardt.
\newblock Measuring massive multitask language understanding.
\newblock \emph{arXiv preprint arXiv:2009.03300}, 2020.

\bibitem[Hewitt(2021)]{hewitt2021initializing}
John Hewitt.
\newblock Initializing new word embeddings for pretrained language models,
  2021.
\newblock URL \url{https:/nlp.stanford.edu/~johnhew//vocab-expansion.html}.

\bibitem[Hoffmann et~al.(2022)Hoffmann, Borgeaud, Mensch, Buchatskaya, Cai,
  Rutherford, Casas, Hendricks, Welbl, Clark, et~al.]{hoffmann2022training}
Jordan Hoffmann, Sebastian Borgeaud, Arthur Mensch, Elena Buchatskaya, Trevor
  Cai, Eliza Rutherford, Diego de~Las Casas, Lisa~Anne Hendricks, Johannes
  Welbl, Aidan Clark, et~al.
\newblock Training compute-optimal large language models.
\newblock \emph{arXiv preprint arXiv:2203.15556}, 2022.

\bibitem[HomebrewResearch(2024)]{InstructionSpeech2024}
HomebrewResearch.
\newblock Instruction speech.
\newblock \emph{Hugging Face Dataset}, 2024.

\bibitem[Hu et~al.(2021)Hu, Shen, Wallis, Allen-Zhu, Li, Wang, Wang, and
  Chen]{hu2021lora}
Edward~J Hu, Yelong Shen, Phillip Wallis, Zeyuan Allen-Zhu, Yuanzhi Li, Shean
  Wang, Lu~Wang, and Weizhu Chen.
\newblock Lora: Low-rank adaptation of large language models.
\newblock \emph{arXiv preprint arXiv:2106.09685}, 2021.

\bibitem[Hu et~al.(2024)Hu, Zhou, Liu, Chen, Hao, Pan, Liu, Li, Sivasankaran,
  Liu, et~al.]{hu2024wavllm}
Shujie Hu, Long Zhou, Shujie Liu, Sanyuan Chen, Hongkun Hao, Jing Pan, Xunying
  Liu, Jinyu Li, Sunit Sivasankaran, Linquan Liu, et~al.
\newblock Wavllm: Towards robust and adaptive speech large language model.
\newblock \emph{arXiv preprint arXiv:2404.00656}, 2024.

\bibitem[HuggingFaceTB(2024)]{huggingfacetb2024everyday}
HuggingFaceTB.
\newblock Everyday-conversations-llama3.1-2k, 2024.
\newblock URL
  \url{https://huggingface.co/datasets/HuggingFaceTB/everyday-conversations-llama3.1-2k}.

\bibitem[Intel(2024)]{intel2024orca}
Intel.
\newblock orca\_dpo\_pairs, 2024.
\newblock URL \url{https://huggingface.co/datasets/Intel/orca_dpo_pairs}.

\bibitem[Lakhotia et~al.(2021)Lakhotia, Kharitonov, Hsu, Adi, Polyak, Bolte,
  Nguyen, Copet, Baevski, Mohamed, et~al.]{lakhotia2021generative}
Kushal Lakhotia, Eugene Kharitonov, Wei-Ning Hsu, Yossi Adi, Adam Polyak,
  Benjamin Bolte, Tu-Anh Nguyen, Jade Copet, Alexei Baevski, Abdelrahman
  Mohamed, et~al.
\newblock On generative spoken language modeling from raw audio.
\newblock \emph{Transactions of the Association for Computational Linguistics},
  9:\penalty0 1336--1354, 2021.

\bibitem[Loshchilov and Hutter(2019)]{adamw}
Ilya Loshchilov and Frank Hutter.
\newblock Decoupled weight decay regularization, 2019.
\newblock URL \url{https://arxiv.org/abs/1711.05101}.

\bibitem[Magpie-Align(2024)]{magpiealign2024magpie}
Magpie-Align.
\newblock Magpie-pro-300k-filtered, 2024.
\newblock URL
  \url{https://huggingface.co/datasets/Magpie-Align/Magpie-Pro-300K-Filtered}.

\bibitem[Maiti et~al.(2024)Maiti, Peng, Choi, Jung, Chang, and
  Watanabe]{maiti2024voxtlm}
Soumi Maiti, Yifan Peng, Shukjae Choi, Jee-weon Jung, Xuankai Chang, and Shinji
  Watanabe.
\newblock Voxtlm: Unified decoder-only models for consolidating speech
  recognition, synthesis and speech, text continuation tasks.
\newblock In \emph{ICASSP 2024-2024 IEEE International Conference on Acoustics,
  Speech and Signal Processing (ICASSP)}, pages 13326--13330. IEEE, 2024.

\bibitem[Microsoft(2024)]{microsoft2024orca}
Microsoft.
\newblock orca-math-word-problems-200k, 2024.
\newblock URL
  \url{https://huggingface.co/datasets/microsoft/orca-math-word-problems-200k}.

\bibitem[Nguyen et~al.(2023)Nguyen, Kharitonov, Copet, Adi, Hsu, Elkahky,
  Tomasello, Algayres, Sagot, Mohamed, et~al.]{nguyen2023generative}
Tu~Anh Nguyen, Eugene Kharitonov, Jade Copet, Yossi Adi, Wei-Ning Hsu, Ali
  Elkahky, Paden Tomasello, Robin Algayres, Benoit Sagot, Abdelrahman Mohamed,
  et~al.
\newblock Generative spoken dialogue language modeling.
\newblock \emph{Transactions of the Association for Computational Linguistics},
  11:\penalty0 250--266, 2023.

\bibitem[nomic ai(2024)]{nomicai2024gpt4all}
nomic ai.
\newblock gpt4all-j-prompt-generations, 2024.
\newblock URL
  \url{https://huggingface.co/datasets/nomic-ai/gpt4all-j-prompt-generations}.

\bibitem[Open-Orca(2024)]{openorca2024oogpt4}
Open-Orca.
\newblock oo-gpt4-200k, 2024.
\newblock URL \url{https://huggingface.co/datasets/Open-Orca/oo-gpt4-200k}.

\bibitem[OpenAI et~al.(2024)OpenAI, Achiam, Adler, Agarwal, Ahmad, Akkaya,
  Aleman, Almeida, Altenschmidt, Altman, Anadkat, Avila, Babuschkin, Balaji,
  Balcom, Baltescu, Bao, Bavarian, Belgum, Bello, Berdine, Bernadett-Shapiro,
  Berner, Bogdonoff, Boiko, Boyd, Brakman, Brockman, Brooks, Brundage, Button,
  Cai, Campbell, Cann, Carey, Carlson, Carmichael, Chan, Chang, Chantzis, Chen,
  Chen, Chen, Chen, Chen, Chess, Cho, Chu, Chung, Cummings, Currier, Dai,
  Decareaux, Degry, Deutsch, Deville, Dhar, Dohan, Dowling, Dunning, Ecoffet,
  Eleti, Eloundou, Farhi, Fedus, Felix, Fishman, Forte, Fulford, Gao, Georges,
  Gibson, Goel, Gogineni, Goh, Gontijo-Lopes, Gordon, Grafstein, Gray, Greene,
  Gross, Gu, Guo, Hallacy, Han, Harris, He, Heaton, Heidecke, Hesse, Hickey,
  Hickey, Hoeschele, Houghton, Hsu, Hu, Hu, Huizinga, Jain, Jain, Jang, Jiang,
  Jiang, Jin, Jin, Jomoto, Jonn, Jun, Kaftan, Łukasz Kaiser, Kamali,
  Kanitscheider, Keskar, Khan, Kilpatrick, Kim, Kim, Kim, Kirchner, Kiros,
  Knight, Kokotajlo, Łukasz Kondraciuk, Kondrich, Konstantinidis, Kosic,
  Krueger, Kuo, Lampe, Lan, Lee, Leike, Leung, Levy, Li, Lim, Lin, Lin, Litwin,
  Lopez, Lowe, Lue, Makanju, Malfacini, Manning, Markov, Markovski, Martin,
  Mayer, Mayne, McGrew, McKinney, McLeavey, McMillan, McNeil, Medina, Mehta,
  Menick, Metz, Mishchenko, Mishkin, Monaco, Morikawa, Mossing, Mu, Murati,
  Murk, Mély, Nair, Nakano, Nayak, Neelakantan, Ngo, Noh, Ouyang, O'Keefe,
  Pachocki, Paino, Palermo, Pantuliano, Parascandolo, Parish, Parparita,
  Passos, Pavlov, Peng, Perelman, de~Avila Belbute~Peres, Petrov,
  de~Oliveira~Pinto, Michael, Pokorny, Pokrass, Pong, Powell, Power, Power,
  Proehl, Puri, Radford, Rae, Ramesh, Raymond, Real, Rimbach, Ross, Rotsted,
  Roussez, Ryder, Saltarelli, Sanders, Santurkar, Sastry, Schmidt, Schnurr,
  Schulman, Selsam, Sheppard, Sherbakov, Shieh, Shoker, Shyam, Sidor, Sigler,
  Simens, Sitkin, Slama, Sohl, Sokolowsky, Song, Staudacher, Such, Summers,
  Sutskever, Tang, Tezak, Thompson, Tillet, Tootoonchian, Tseng, Tuggle,
  Turley, Tworek, Uribe, Vallone, Vijayvergiya, Voss, Wainwright, Wang, Wang,
  Wang, Ward, Wei, Weinmann, Welihinda, Welinder, Weng, Weng, Wiethoff,
  Willner, Winter, Wolrich, Wong, Workman, Wu, Wu, Wu, Xiao, Xu, Yoo, Yu, Yuan,
  Zaremba, Zellers, Zhang, Zhang, Zhao, Zheng, Zhuang, Zhuk, and
  Zoph]{openai2024gpt4technicalreport}
OpenAI, Josh Achiam, Steven Adler, Sandhini Agarwal, Lama Ahmad, Ilge Akkaya,
  Florencia~Leoni Aleman, Diogo Almeida, Janko Altenschmidt, Sam Altman,
  Shyamal Anadkat, Red Avila, Igor Babuschkin, Suchir Balaji, Valerie Balcom,
  Paul Baltescu, Haiming Bao, Mohammad Bavarian, Jeff Belgum, Irwan Bello, Jake
  Berdine, Gabriel Bernadett-Shapiro, Christopher Berner, Lenny Bogdonoff, Oleg
  Boiko, Madelaine Boyd, Anna-Luisa Brakman, Greg Brockman, Tim Brooks, Miles
  Brundage, Kevin Button, Trevor Cai, Rosie Campbell, Andrew Cann, Brittany
  Carey, Chelsea Carlson, Rory Carmichael, Brooke Chan, Che Chang, Fotis
  Chantzis, Derek Chen, Sully Chen, Ruby Chen, Jason Chen, Mark Chen, Ben
  Chess, Chester Cho, Casey Chu, Hyung~Won Chung, Dave Cummings, Jeremiah
  Currier, Yunxing Dai, Cory Decareaux, Thomas Degry, Noah Deutsch, Damien
  Deville, Arka Dhar, David Dohan, Steve Dowling, Sheila Dunning, Adrien
  Ecoffet, Atty Eleti, Tyna Eloundou, David Farhi, Liam Fedus, Niko Felix,
  Simón~Posada Fishman, Juston Forte, Isabella Fulford, Leo Gao, Elie Georges,
  Christian Gibson, Vik Goel, Tarun Gogineni, Gabriel Goh, Rapha Gontijo-Lopes,
  Jonathan Gordon, Morgan Grafstein, Scott Gray, Ryan Greene, Joshua Gross,
  Shixiang~Shane Gu, Yufei Guo, Chris Hallacy, Jesse Han, Jeff Harris, Yuchen
  He, Mike Heaton, Johannes Heidecke, Chris Hesse, Alan Hickey, Wade Hickey,
  Peter Hoeschele, Brandon Houghton, Kenny Hsu, Shengli Hu, Xin Hu, Joost
  Huizinga, Shantanu Jain, Shawn Jain, Joanne Jang, Angela Jiang, Roger Jiang,
  Haozhun Jin, Denny Jin, Shino Jomoto, Billie Jonn, Heewoo Jun, Tomer Kaftan,
  Łukasz Kaiser, Ali Kamali, Ingmar Kanitscheider, Nitish~Shirish Keskar,
  Tabarak Khan, Logan Kilpatrick, Jong~Wook Kim, Christina Kim, Yongjik Kim,
  Jan~Hendrik Kirchner, Jamie Kiros, Matt Knight, Daniel Kokotajlo, Łukasz
  Kondraciuk, Andrew Kondrich, Aris Konstantinidis, Kyle Kosic, Gretchen
  Krueger, Vishal Kuo, Michael Lampe, Ikai Lan, Teddy Lee, Jan Leike, Jade
  Leung, Daniel Levy, Chak~Ming Li, Rachel Lim, Molly Lin, Stephanie Lin,
  Mateusz Litwin, Theresa Lopez, Ryan Lowe, Patricia Lue, Anna Makanju, Kim
  Malfacini, Sam Manning, Todor Markov, Yaniv Markovski, Bianca Martin, Katie
  Mayer, Andrew Mayne, Bob McGrew, Scott~Mayer McKinney, Christine McLeavey,
  Paul McMillan, Jake McNeil, David Medina, Aalok Mehta, Jacob Menick, Luke
  Metz, Andrey Mishchenko, Pamela Mishkin, Vinnie Monaco, Evan Morikawa, Daniel
  Mossing, Tong Mu, Mira Murati, Oleg Murk, David Mély, Ashvin Nair, Reiichiro
  Nakano, Rajeev Nayak, Arvind Neelakantan, Richard Ngo, Hyeonwoo Noh, Long
  Ouyang, Cullen O'Keefe, Jakub Pachocki, Alex Paino, Joe Palermo, Ashley
  Pantuliano, Giambattista Parascandolo, Joel Parish, Emy Parparita, Alex
  Passos, Mikhail Pavlov, Andrew Peng, Adam Perelman, Filipe de~Avila
  Belbute~Peres, Michael Petrov, Henrique~Ponde de~Oliveira~Pinto, Michael,
  Pokorny, Michelle Pokrass, Vitchyr~H. Pong, Tolly Powell, Alethea Power,
  Boris Power, Elizabeth Proehl, Raul Puri, Alec Radford, Jack Rae, Aditya
  Ramesh, Cameron Raymond, Francis Real, Kendra Rimbach, Carl Ross, Bob
  Rotsted, Henri Roussez, Nick Ryder, Mario Saltarelli, Ted Sanders, Shibani
  Santurkar, Girish Sastry, Heather Schmidt, David Schnurr, John Schulman,
  Daniel Selsam, Kyla Sheppard, Toki Sherbakov, Jessica Shieh, Sarah Shoker,
  Pranav Shyam, Szymon Sidor, Eric Sigler, Maddie Simens, Jordan Sitkin,
  Katarina Slama, Ian Sohl, Benjamin Sokolowsky, Yang Song, Natalie Staudacher,
  Felipe~Petroski Such, Natalie Summers, Ilya Sutskever, Jie Tang, Nikolas
  Tezak, Madeleine~B. Thompson, Phil Tillet, Amin Tootoonchian, Elizabeth
  Tseng, Preston Tuggle, Nick Turley, Jerry Tworek, Juan Felipe~Cerón Uribe,
  Andrea Vallone, Arun Vijayvergiya, Chelsea Voss, Carroll Wainwright,
  Justin~Jay Wang, Alvin Wang, Ben Wang, Jonathan Ward, Jason Wei, CJ~Weinmann,
  Akila Welihinda, Peter Welinder, Jiayi Weng, Lilian Weng, Matt Wiethoff, Dave
  Willner, Clemens Winter, Samuel Wolrich, Hannah Wong, Lauren Workman, Sherwin
  Wu, Jeff Wu, Michael Wu, Kai Xiao, Tao Xu, Sarah Yoo, Kevin Yu, Qiming Yuan,
  Wojciech Zaremba, Rowan Zellers, Chong Zhang, Marvin Zhang, Shengjia Zhao,
  Tianhao Zheng, Juntang Zhuang, William Zhuk, and Barret Zoph.
\newblock Gpt-4 technical report, 2024.
\newblock URL \url{https://arxiv.org/abs/2303.08774}.

\bibitem[Panayotov et~al.(2015)Panayotov, Chen, Povey, and
  Khudanpur]{panayotov2015librispeech}
Vassil Panayotov, Guoguo Chen, Daniel Povey, and Sanjeev Khudanpur.
\newblock Librispeech: an asr corpus based on public domain audio books.
\newblock In \emph{2015 IEEE International Conference on Acoustics, Speech and
  Signal Processing (ICASSP)}. IEEE, 2015.

\bibitem[Paszke et~al.(2019)Paszke, Gross, Massa, Lerer, Bradbury, Chanan,
  Killeen, Lin, Gimelshein, Antiga, Desmaison, Kopf, Yang, DeVito, Raison,
  Tejani, Chilamkurthy, Steiner, Fang, Bai, and Chintala]{paszke2019pytorch}
Adam Paszke, Sam Gross, Francisco Massa, Adam Lerer, James Bradbury, Gregory
  Chanan, Trevor Killeen, Zeming Lin, Natalia Gimelshein, Luca Antiga, Alban
  Desmaison, Andreas Kopf, Edward Yang, Zachary DeVito, Martin Raison, Alykhan
  Tejani, Sasank Chilamkurthy, Benoit Steiner, Lu~Fang, Junjie Bai, and Soumith
  Chintala.
\newblock Pytorch: An imperative style, high-performance deep learning library.
\newblock In \emph{Advances in Neural Information Processing Systems 32}, pages
  8024--8035. Curran Associates, Inc., 2019.
\newblock URL
  \url{http://papers.neurips.cc/paper/9015-pytorch-an-imperative-style-high-performance-deep-learning-library.pdf}.

\bibitem[PJMixers(2024)]{pjmixers2024math}
PJMixers.
\newblock Math-multiturn-10k-sharegpt, 2024.
\newblock URL
  \url{https://huggingface.co/datasets/PJMixers/Math-Multiturn-10K-ShareGPT}.

\bibitem[Pratap et~al.(2020)Pratap, Xu, Sriram, Synnaeve, and
  Collobert]{Pratap2020MLSAL}
Vineel Pratap, Qiantong Xu, Anuroop Sriram, Gabriel Synnaeve, and Ronan
  Collobert.
\newblock Mls: A large-scale multilingual dataset for speech research.
\newblock \emph{ArXiv}, abs/2012.03411, 2020.

\bibitem[qiaojin(2024)]{qiaojin2024pubmedqa}
qiaojin.
\newblock Pubmedqa, 2024.
\newblock URL \url{https://huggingface.co/datasets/qiaojin/PubMedQA}.

\bibitem[Radford et~al.(2019)Radford, Wu, Child, Luan, Amodei, Sutskever,
  et~al.]{radford2019language}
Alec Radford, Jeffrey Wu, Rewon Child, David Luan, Dario Amodei, Ilya
  Sutskever, et~al.
\newblock Language models are unsupervised multitask learners.
\newblock \emph{OpenAI blog}, 1\penalty0 (8):\penalty0 9, 2019.

\bibitem[Radford et~al.(2022)Radford, Kim, Xu, Brockman, McLeavey, and
  Sutskever]{Whisper}
Alec Radford, Jong~Wook Kim, Tao Xu, Greg Brockman, Christine McLeavey, and
  Ilya Sutskever.
\newblock Robust speech recognition via large-scale weak supervision.
\newblock 2022.
\newblock URL \url{https://arxiv.org/abs/2212.04356}.

\bibitem[Raffel et~al.(2020)Raffel, Shazeer, Roberts, Lee, Narang, Matena,
  Zhou, Li, and Liu]{raffel2020exploring}
Colin Raffel, Noam Shazeer, Adam Roberts, Katherine Lee, Sharan Narang, Michael
  Matena, Yanqi Zhou, Wei Li, and Peter~J Liu.
\newblock Exploring the limits of transfer learning with a unified text-to-text
  transformer.
\newblock \emph{Journal of machine learning research}, 21\penalty0
  (140):\penalty0 1--67, 2020.

\bibitem[Ramesh et~al.(2022)Ramesh, Dhariwal, Nichol, Chu, and
  Chen]{ramesh2022hierarchical}
Aditya Ramesh, Prafulla Dhariwal, Alex Nichol, Casey Chu, and Mark Chen.
\newblock Hierarchical text-conditional image generation with clip latents.
\newblock \emph{arXiv preprint arXiv:2204.06125}, 1\penalty0 (2):\penalty0 3,
  2022.

\bibitem[Rein et~al.(2023)Rein, Hou, Stickland, Petty, Pang, Dirani, Michael,
  and Bowman]{rein2023gpqa}
David Rein, Betty~Li Hou, Asa~Cooper Stickland, Jackson Petty, Richard~Yuanzhe
  Pang, Julien Dirani, Julian Michael, and Samuel~R Bowman.
\newblock Gpqa: A graduate-level google-proof q\&a benchmark.
\newblock \emph{arXiv preprint arXiv:2311.12022}, 2023.

\bibitem[routellm(2024)]{routellm2024gpt4}
routellm.
\newblock gpt4\_dataset, 2024.
\newblock URL \url{https://huggingface.co/datasets/routellm/gpt4_dataset}.

\bibitem[Rubenstein et~al.(2023)Rubenstein, Asawaroengchai, Nguyen, Bapna,
  Borsos, Quitry, Chen, Badawy, Han, Kharitonov,
  et~al.]{rubenstein2023audiopalm}
Paul~K Rubenstein, Chulayuth Asawaroengchai, Duc~Dung Nguyen, Ankur Bapna,
  Zal{\'a}n Borsos, F{\'e}lix de~Chaumont Quitry, Peter Chen, Dalia~El Badawy,
  Wei Han, Eugene Kharitonov, et~al.
\newblock Audiopalm: A large language model that can speak and listen.
\newblock \emph{arXiv preprint arXiv:2306.12925}, 2023.

\bibitem[Shu et~al.(2023)Shu, Dong, Chen, Huang, Zhang, Shi, Xiang, and
  Shi]{shu2023llasm}
Yu~Shu, Siwei Dong, Guangyao Chen, Wenhao Huang, Ruihua Zhang, Daochen Shi,
  Qiqi Xiang, and Yemin Shi.
\newblock Llasm: Large language and speech model.
\newblock \emph{arXiv preprint arXiv:2308.15930}, 2023.

\bibitem[Tang et~al.(2023)Tang, Yu, Sun, Chen, Tan, Li, Lu, Ma, and
  Zhang]{tang2023salmonn}
Changli Tang, Wenyi Yu, Guangzhi Sun, Xianzhao Chen, Tian Tan, Wei Li, Lu~Lu,
  Zejun Ma, and Chao Zhang.
\newblock Salmonn: Towards generic hearing abilities for large language models.
\newblock \emph{arXiv preprint arXiv:2310.13289}, 2023.

\bibitem[Team(2024{\natexlab{a}})]{team2024chameleon}
Chameleon Team.
\newblock Chameleon: Mixed-modal early-fusion foundation models.
\newblock \emph{arXiv preprint arXiv:2405.09818}, 2024{\natexlab{a}}.

\bibitem[Team(2024{\natexlab{b}})]{qwen2.5}
Qwen Team.
\newblock Qwen2.5: A party of foundation models, September 2024{\natexlab{b}}.
\newblock URL \url{https://qwenlm.github.io/blog/qwen2.5/}.

\bibitem[torchtune maintainers and contributors(2024)]{torchtune}
torchtune maintainers and contributors.
\newblock torchtune: Pytorch's finetuning library, 2024.
\newblock URL \url{https//github.com/pytorch/torchtune}.

\bibitem[Touvron et~al.(2023)Touvron, Lavril, Izacard, Martinet, Lachaux,
  Lacroix, Rozi{\`e}re, Goyal, Hambro, Azhar, et~al.]{touvron2023llama}
Hugo Touvron, Thibaut Lavril, Gautier Izacard, Xavier Martinet, Marie-Anne
  Lachaux, Timoth{\'e}e Lacroix, Baptiste Rozi{\`e}re, Naman Goyal, Eric
  Hambro, Faisal Azhar, et~al.
\newblock Llama: Open and efficient foundation language models.
\newblock \emph{arXiv preprint arXiv:2302.13971}, 2023.

\bibitem[Undi95(2024)]{undi952024capybara}
Undi95.
\newblock Capybara-sharegpt, 2024.
\newblock URL \url{https://huggingface.co/datasets/Undi95/Capybara-ShareGPT}.

\bibitem[Wadekar et~al.(2024)Wadekar, Chaurasia, Chadha, and
  Culurciello]{wadekar2024evolution}
Shakti~N Wadekar, Abhishek Chaurasia, Aman Chadha, and Eugenio Culurciello.
\newblock The evolution of multimodal model architectures.
\newblock \emph{arXiv preprint arXiv:2405.17927}, 2024.

\bibitem[Wang et~al.(2024)Wang, Zou, Lin, Sun, Liu, Zhang, Liu, Aw, and
  Chen]{wang2024audiobench}
Bin Wang, Xunlong Zou, Geyu Lin, Shuo Sun, Zhuohan Liu, Wenyu Zhang, Zhengyuan
  Liu, AiTi Aw, and Nancy~F Chen.
\newblock Audiobench: A universal benchmark for audio large language models.
\newblock \emph{arXiv preprint arXiv:2406.16020}, 2024.

\bibitem[Wang et~al.(2023{\natexlab{a}})Wang, Chen, Wu, Zhang, Zhou, Liu, Chen,
  Liu, Wang, Li, et~al.]{wang2023neural}
Chengyi Wang, Sanyuan Chen, Yu~Wu, Ziqiang Zhang, Long Zhou, Shujie Liu, Zhuo
  Chen, Yanqing Liu, Huaming Wang, Jinyu Li, et~al.
\newblock Neural codec language models are zero-shot text to speech
  synthesizers.
\newblock \emph{arXiv preprint arXiv:2301.02111}, 2023{\natexlab{a}}.

\bibitem[Wang et~al.(2023{\natexlab{b}})Wang, Zhou, Zhang, Wu, Liu, Gaur, Chen,
  Li, and Wei]{wang2023viola}
Tianrui Wang, Long Zhou, Ziqiang Zhang, Yu~Wu, Shujie Liu, Yashesh Gaur, Zhuo
  Chen, Jinyu Li, and Furu Wei.
\newblock Viola: Unified codec language models for speech recognition,
  synthesis, and translation.
\newblock \emph{arXiv preprint arXiv:2305.16107}, 2023{\natexlab{b}}.

\bibitem[Xu et~al.(2024)Xu, Jiang, Niu, Deng, Poovendran, Choi, and
  Lin]{xu2024magpie}
Zhangchen Xu, Fengqing Jiang, Luyao Niu, Yuntian Deng, Radha Poovendran, Yejin
  Choi, and Bill~Yuchen Lin.
\newblock Magpie: Alignment data synthesis from scratch by prompting aligned
  llms with nothing.
\newblock \emph{arXiv preprint arXiv:2406.08464}, 2024.

\bibitem[Zhan et~al.(2024)Zhan, Dai, Ye, Zhou, Zhang, Liu, Zhang, Yuan, Zhang,
  Li, et~al.]{zhan2024anygpt}
Jun Zhan, Junqi Dai, Jiasheng Ye, Yunhua Zhou, Dong Zhang, Zhigeng Liu, Xin
  Zhang, Ruibin Yuan, Ge~Zhang, Linyang Li, et~al.
\newblock Anygpt: Unified multimodal llm with discrete sequence modeling.
\newblock \emph{arXiv preprint arXiv:2402.12226}, 2024.

\bibitem[Zhang et~al.(2023)Zhang, Zhang, Li, Zhou, and
  Qiu]{zhang2023speechtokenizer}
Xin Zhang, Dong Zhang, Shimin Li, Yaqian Zhou, and Xipeng Qiu.
\newblock Speechtokenizer: Unified speech tokenizer for speech large language
  models.
\newblock \emph{arXiv preprint arXiv:2308.16692}, 2023.

\end{thebibliography}
